# Exploiting First-Order Regression in Inductive Policy Selection

**Charles Gretton** and **Sylvie Thiébaux**
National ICT Australia &
Computer Sciences Laboratory
The Australian National University
Canberra, ACT 0200, Australia
{charlesg,thiebaux}@csl.anu.edu.au

**Abstract**

We consider the problem of computing optimal generalised policies for relational Markov decision processes. We describe an approach combining some of the benefits of purely inductive techniques with those of symbolic dynamic programming methods. The latter reason about the optimal value function using first-order decision-theoretic regression and formula rewriting, while the former, when provided with a suitable hypotheses language, are capable of generalising value functions or policies for small instances. Our idea is to use reasoning and in particular classical first-order regression to automatically generate a hypotheses language dedicated to the domain at hand, which is then used as input by an inductive solver. This approach avoids the more complex reasoning of symbolic dynamic programming while focusing the inductive solver's attention on concepts that are specifically relevant to the optimal value function for the domain considered.

## 1 INTRODUCTION

Planning domains often exhibit a strong relational structure and are therefore traditionally represented using first-order languages supporting the declaration of objects and relations over them as well as the use of quantification over objects [12]. Although Markov decision processes (MDPs) are now widely accepted as the preferred model for decision-theoretic planning, state of the art MDP algorithms operate on either state-based or propositionally factored representations [14, 2, 15, 8], thereby failing to exploit the relational structure of planning domains. Due to the size of these representations, such approaches do not scale very well as the number of objects increases. Furthermore, they do little in the way of addressing the long-standing goal of generating *generalised* policies that are applicable to an arbitrary number of objects. Instead, MDP planners usually replan from scratch when computing a policy for an instance with marginally more or fewer states.

Recent research on *relational MDPs* has started to address these issues. Relational approaches fall mainly into two classes. Approaches in the first class extend dynamic programming methods to operate directly on first-order domain and value function descriptions [4]. The $n$-stage-to-go value function, represented as a mapping from a set of first-order formulae partitioning the state space to the reals, is obtained by pure logical reasoning. This involves in particular reasoning about the domain dynamics using a first-order version of *decision-theoretic regression* [3], and reasoning about maximisation using formula rewriting. While this approach is theoretically attractive, a difficult challenge is to implement effective formula simplification rules and theorem proving techniques to keep the formulae consistent and of manageable size. We are unaware of existing implementations that successfully address this challenge.

Approaches in the second class avoid those problems by employing inductive learning techniques: they generalise good policies (or value functions) for instances with a small number of objects to get a useful generalised policy [16, 7, 18, 23, 19]. In order to address domains whose small instances are not representative of the general case, they can be made to induce policies which are likely to generalise well if, for instance, training data in the form of a policy trajectory, a list of propositional state action pairs, generated by approximate policy iteration is used [10, 11]. Inductive learning proposals do not reason about the domain dynamics beyond generation of the training data. In contrast to the dynamic programming approach above these do not explicitly seek optimality (or in some cases correctness). This feature is motivated by the fact that domains arise where no practical representation of the optimal generalised value function or policy exists.

To keep the search space manageable, inductive approaches require a suitable *hypotheses language*, sufficiently rich to describe the control strategies of interest without wasting the learner's time on irrelevant planning concepts. This can take the form of support predicates that express key features of the domain in terms of the basic relations (e.g. "above" and "in-position" in blocks world) [16, 7], or that of a domain-independent language bias from which the im-portant features can be discovered from scratch – for example a concept language based on description or taxonomic logics appears to be well suited to blocks world and logistics benchmarks [18, 23, 10]. The main weakness of inductive approaches is their reliance on a suitable hypotheses language. One can also question the fact that they never explicitly reason about the known domain dynamics or exploit it beyond the generation of training data. Although this makes them more flexible and practical than the decision-theoretic regression approach, this may be seen as a failure to exploit useful information.

In this paper, we consider the problem of computing optimal generalised policies given a first-order domain and reward descriptions. We investigate an approach aimed at combining some of the strengths of dynamic programming and inductive techniques. Our idea is to automatically generate a suitable hypotheses language for the domain at hand, by reasoning about the dynamics of this domain using first-order regression. This language is guaranteed to cover all concepts relevant to the optimal $n$-stage-to-go value function for a given $n$, and can be used as input by any inductive solver. More explicitly, we repeatedly apply *classical* first-order regression (see e.g. [21]) to the first-order formulae involved in the reward description to generate candidate formulae for inclusion in the $n$-stage-to-go generalised value function. The inductive solver selects among those formulae to build a decision tree generalising small value functions generated by a state of the art MDP solver. Because we avoid much of the most expensive reasoning performed by dynamic programming approaches, we are able to retain acceptable performance. Because our hypotheses language is targeted at the domain of interest, we are often able to obtain optimal generalised policies using very few training examples.

The paper is organised as follows. We start with background material on MDPs, relational MDPs, first-order regression, and previous approaches. We follow by a description of our approach, together with a discussion of its strengths and weaknesses. We then present experimental results before concluding with some remarks about related and future work.

## 2 BACKGROUND

### 2.1 MDPs

We take a Markov decision process to be a 4-tuple $\langle \mathcal{E}, \mathcal{A}, \Pr, \mathcal{R} \rangle$, where $\mathcal{E}$ is a possibly infinite set of fully observable states, $\mathcal{A}$ is a possibly infinite set of (ground) actions ($\mathcal{A}(e)$ denotes the subset of actions applicable in $e \in \mathcal{E}$), $\{\Pr(e, a, \bullet) \mid e \in \mathcal{E}, a \in \mathcal{A}(e)\}$ is a family of probability distributions over $\mathcal{E}$ such that $\Pr(e, a, e')$ is the probability of being in state $e'$ after performing action $a$ in state $e$, and $\mathcal{R} : \mathcal{E} \to \mathbb{R}$ is a reward function such that $\mathcal{R}(e)$ is the immediate reward for being in state $e$. A stationary policy for an MDP is a function $\pi : \mathcal{E} \mapsto \mathcal{A}$, such that $\pi(e) \in \mathcal{A}(e)$ is the action to be executed in state $e$. The value $V_\pi(e)$ of state $e$ under the policy is the sum of the expected future rewards, discounted by how far into the future they occur:

$$V_\pi(e) = \lim_{n \to \infty} \mathsf{E}\left[\sum_{t=0}^{n} \beta^t \mathcal{R}(e_t) \mid \pi, e_0 = e\right]$$

where $0 \leq \beta < 1$ is the discounting factor controlling the contribution of distant rewards and $e_t$ is the state at time $t$. Policy $\pi$ is optimal iff $V_\pi(e) \geq V_{\pi'}(e)$ for all $e \in \mathcal{E}$ and all policies $\pi'$.

### 2.2 RELATIONAL MDPs

While the above state-based definition of MDPs is suitable as a general mathematical model, it fails to emphasise the relational structure of planning problems. For this reason, recent research has focused on relational MDPs, which make this structure explicit and open the way to algorithms capable of exploiting it. Under the relational model, MDPs are often represented using a first-order formalism supporting relations, functions, and quantification over objects. Some of the most famous formalisms used for that purpose are first order probabilistic STRIPS variants [1, 9, 23, 24] and the situation calculus [21, 4]. Our presentation uses the situation calculus, as we believe it provides clear logical foundations for our approach.

The situation calculus has 3 disjoint sorts: actions, situations and objects. The alphabet includes variables of each sort, function and predicate symbols of sort $object^n \to object$ and $object^n$, respectively, used to denote situation-independent functions and relations, as well as the usual connectives and quantifiers $\neg, \wedge, \exists$ with the usual abbreviations $\vee, \to, \forall$, etc. Other elements of the language include the following.

*Actions* are first-order terms built from an action function symbol of sort $object^n \to action$ and its arguments. For instance in the following, $move(x, y)$ denotes the action of moving object x onto object y. When the arguments are ground, we sometimes speak of a *ground* action. In what follows, we shall only make the distinction between actions and ground actions when that distinction matters.

*Situation* terms are built using two symbols: a constant symbol $S_0$ denoting the initial situation, and the function symbol $do : action \times situation \to situation$, with the interpretation that $do(a, s)$ denotes the situation resulting from performing deterministic action $a$ in situation $s$.

*Relations* whose truth values vary from situation to situation are built using predicate symbols of sort $object^n \times situation$ called relational fluent symbols. For instance $On(x, y, s)$ is a relational fluent meaning that object $x$ is on object $y$ in situation $s$.

Additionally, there is a predicate symbol $poss$ of sort $action \times situation$. The intended interpretation of $poss(a, s)$ is that it is possible to perform deterministic action $a$ in situation $s$.

The situation calculus views stochastic actions as probability distributions over deterministic actions. Executing a stochastic action amounts to letting "nature" choose which deterministic action will be executed, this choice being governed by given probabilities. Describing stochastic actions requires (1) a predicate symbol $choice : action \times action$, where $choice(da, sa)$ denotes that executing deterministic action $da$ is a possible nature choice when executing stochastic action $sa$, and (2) a function symbol $prob : action \times action \times situation \to \mathbb{R}$, where $prob(da, sa, s)$ denotes the probability of that choice in situation $s$.

Finally, function symbol $R : situation \to \mathbb{R}$ is used to denote the immediate reward received in a situation.[1]

As in [4], we use the notion of a *state formula*, $f(\vec{x}, s)$, whose only free variables are non-situation variables $\vec{x}$ and situation variable $s$, and in which no other situation term occurs.[2] Intuitively, a state formula $f(\vec{x}, s)$ only refers to properties of situation $s$. We say that an MDP state *models* a state formula whose only free variable is situation $s$ iff the properties of $s$ described by the formula hold in the state.[3] A set of state formulae $\{f_i(\vec{x}, s)\}$ *partitions* the state space iff $\models \forall \vec{x} \forall s (\vee_i f_i(\vec{x}, s))$ and for all $i$ and all $j \neq i$ $\models \forall \vec{x} \forall s (f_i(\vec{x}, s) \to \neg f_j(\vec{x}, s))$.

Modelling a relational MDP in the situation calculus involves writing the following axioms.

1. *Reward axiom:* rewards in the current situation are conveniently expressed as a statement of the form: $R(s) = case[\rho_1(s), r_1; \ldots; \rho_n(s), r_n]$, where the $r_i$s are reals, the $\rho_i$s are state formulae partitioning the state space, and where the notation $t = case[f_1, t_1; \ldots; f_n, t_n]$ abbreviates $\vee_{i=1}^n (f_i \wedge t = t_i)$. For instance, consider a blocks world domain where we get rewarded when all blocks are in their goal position, then:

$R(s) \equiv$
$\quad case[\ \forall b1\ \forall b2\ (OnG(b1, b2) \to On(b1, b2, s)), 100.0;$
$\quad\quad \exists b1\ \exists b2\ (OnG(b1, b2) \wedge \neg On(b1, b2, s)), 0.0]$

where $OnG(b1, b2)$ is a situation-independent relation representing the goal configuration.

2. *Nature's choice and probability axioms:* for each stochastic action $A(\vec{x})$, we must specify the deterministic actions $D_1(\vec{x}), \ldots, D_k(\vec{x})$ available for nature to choose from, via the axiom: $choice(a, A(\vec{x})) \equiv \vee_{j=1}^k (a = D_j(\vec{x}))$. We must also define the probabilities of the choices in the current situation $s$, using axioms of the form: $prob(D_j(\vec{x}), A(\vec{x}), s) = case[\phi_j^1(\vec{x}, s), p_j^1; \ldots; \phi_j^m(\vec{x}, s), p_j^m]$, where the $\phi_j^i$s are state formulae partitioning the state space, and the $p_j^i$s are probabilities. For instance, suppose that in our blocks world domain, the $move(x, y)$ action is stochastic and sometimes behaves like the deterministic $moveS(x, y)$ action which succeeds in moving $x$ to $y$, and otherwise behaves like the deterministic $moveF(x, y)$ action which fails to change anything. Suppose furthermore that the probability of a successful move is 0.9 when the weather is fine, and 0.7 when it is rainy, we get the following axioms:

$choice(a, move(b1, b2)) \equiv$
$\quad a = moveS(b1, b2) \vee a = moveF(b1, b2)$
$prob(moveS(b1, b2), move(b1, b2), s) =$
$\quad case[Rain(s), 0.7; \neg Rain(s), 0.9]$
$prob(moveF(b1, b2), move(b1, b2), s) =$
$\quad case[Rain(s), 0.3; \neg Rain(s), 0.1]$

3. *Action precondition axioms:* for each deterministic action $A(\vec{x})$, we need to write one axiom of the form: $poss(A(\vec{x}), s) \equiv \Psi_A(\vec{x}, s)$, where $\Psi_A(\vec{x}, s)$ is a state formula characterising the preconditions of the action. E.g:

$poss(moveS(b1, b2), s) \equiv poss(moveF(b1, b2), s) \equiv$
$\quad b1 \neq table \wedge b1 \neq b2 \wedge \not\exists b3\ On(b3, b1, s) \wedge$
$\quad (b2 = table \vee \not\exists b3\ On(b3, b2, s))$

4. *Successor states axioms:* they are the means by which the deterministic dynamics of the system is described. For each relational fluent $F(\vec{x}, s)$, there is one axiom of the form: $F(\vec{x}, do(a, s)) \equiv \Phi_F(\vec{x}, a, s)$, where $\Phi_F(\vec{x}, a, s)$ is a state formula characterising the truth value of $F$ in the situation resulting from performing $a$ in $s$. For instance:

$On(b1, b2, do(a, s)) \equiv a = moveS(b1, b2) \vee$
$\quad (On(b1, b2, s) \wedge \not\exists b3\ (b3 \neq b2 \wedge a = moveS(b1, b3)))$
$Rain(do(a, s)) \equiv Rain(s)$

5. Finally, for each pair of distinct actions, we need a unique name axiom of the form $\forall \vec{x} \forall \vec{y}\ A(\vec{x}) \neq B(\vec{y})$, and for each action, there is an "all-different" axiom of the form $\forall \vec{x} \forall \vec{y}\ A(\vec{x}) = A(\vec{y}) \leftrightarrow \vec{x} = \vec{y}$.

This completes our description of relational MDPs in the situation calculus framework. Let us emphasise once more that (1) the modelling retains the classical situation calculus machinery for deterministic domains –stochastic actions only appear in an extra layer on top of this machinery– and that (2), the axioms do not restrict the domain to a pre-specified or even finite set of objects, which is why a solution for an MDP axiomatised this way is a generalised policy applying to an arbitrary object universe. Generalised value functions and policies can conveniently be represented in the situation calculus as a case statement involving state formulae partitioning the state space and real or action terms, respectively.

---

[1] This can be extended to depend on the current action.

[2] In [21], such formulae are said to be *uniform* in $s$. Here we also assume that state formulae do not contain statements involving predicates $poss$ and $choice$, and functions $prob$ and $R$.

[3] We would like to be able to say that MDP states are first-order models of state formulae, but this is not strictly accurate because of the presence of situation variables in fluents. We would have to strip out state variables and reduce the arity of relations. Spelling this out formally would be a waste of space.

### 2.3 FIRST-ORDER REGRESSION

As with many action formalisms, *regression* is the corner stone of reasoning about the dynamics of a deterministic domain in the situation calculus. As usual, the regression of a formula $f$ through a deterministic action $\alpha$ is a formula that holds before $\alpha$ is executed if and only if $f$ holds after the execution. In the situation calculus, regression takes the following form. Consider a state formula $f(\vec{x}, s)$ and an action term $A(\vec{y})$. $f$ holds of the situation $do(A(\vec{y}), \sigma)$ resulting from executing $A(\vec{y})$ in $\sigma$ iff $\Psi_A(\vec{y}, \sigma) \wedge Regr(f(\vec{x}, do(A(\vec{y}), \sigma)))$ holds, where $Regr$ is defined as follows:

- $Regr(F(\vec{t}, do(\alpha, \sigma))) = \Phi_F(\vec{t}, \alpha, \sigma)$ where $F(\vec{x}, do(a, s)) \equiv \Phi_F(\vec{x}, a, s)$ is a successor state axiom[4]
- $Regr(\neg f) = \neg Regr(f)$
- $Regr(f_1 \wedge f_2) = Regr(f_1) \wedge Regr(f_2)$
- $Regr(\exists x\ f) = \exists x\ Regr(f)$
- $Regr(f) = f$ in all other cases

E.g., regressing the formula $\forall b1 \forall b2 (OnG(b1, b2) \rightarrow On(b1, b2, s))$ in our reward description with action $moveS(x, y)$ yields:

$x \neq table \wedge x \neq y \wedge \not\exists b3\ On(b3, x, s)\ \wedge$
$(y = table \vee \not\exists b3\ On(b3, y, s)) \wedge$
$\forall b1\ \forall b2\ (OnG(b1, b2) \rightarrow ((x = b1 \wedge y = b2) \vee$
$(On(b1, b2, s) \wedge (y \neq b2 \rightarrow x \neq b1))))$

meaning that for the goal to be achieved after the move, the move must be executable and for any subgoal $OnG(b1, b2)$, either the move achieves it, or the subgoal was already true and the move does not destroy it.

### 2.4 FIRST-ORDER DYNAMIC PROGRAMMING

One of the very first approaches to solving relational MDPs is first-order symbolic dynamic programming [4]. This is a value iteration approach which directly operates on the symbolic representation of the generalised value function as a case statement. It relies on *first-order decision-theoretic regression*, an extension of regression, as defined above, to stochastic actions. Given a stochastic action $A(\vec{x})$ and the logical description of the generalised $n$-stage-to-go value function $V^n(s)$, first-order decision-theoretic regression is able to compute the logical description of the generalised $n + 1$-stage-to-go $Q$ function $Q^{n+1}(A(\vec{x}), s)$. At each value iteration step, the $Q^{n+1}$ functions are computed for the various actions, and a formula is built expressing

---

[4]Quantifiers in $\Phi_F(\vec{x}, a, s)$ should have their quantified variable renamed as needed to make it different from the free variables in $F(\vec{t}, do(\alpha, \sigma))$. A similar remark applies to quantifiers in $\Psi_A(\vec{x}, s)$ when substituting as above $\vec{y}$ for $\vec{x}$ and $\sigma$ for $s$.

that $V^{n+1}$ is the maximum of the $Q^{n+1}$ functions over the actions.

A drawback of first-order dynamic programming is the practicality of retaining manageable case expressions of value functions: the length and number of formulae included in the case statements rapidly becomes impractically large. This is especially exacerbated by the symbolic maximisation of the $Q$ functions which requires combining already complex formulae obtained through first-order decision-theoretic regression. Another complication is the need for detecting inconsistent expressions which may form, e.g., as a result of such combinations.

By implementing logical simplification rules and enlisting the theorem prover Otter [20] we were able to significantly reduce case bloat and eliminate some forms of redundancies and contradictions. Unfortunately all this comes at a cost. Essentially we find that dynamic programming remains impractical for as little as 3 or 4 value iteration steps for standard planning benchmarks such as blocks world or logistics.

## 3 THE APPROACH

Having examined the shortcomings of the dynamic programming approach, we now seek to extract and apply its essence that is reasoning, using first-order regression, in a different context, namely that of inductive learning. Briefly, our approach uses *classical* first-order regression, as defined in Section 2.3, to generate a hypotheses language for the inductive learner. This language consists of state formulae from which the inductive learner selects to build a decision-tree generalising small instances generated by a conventional MDP solver.

### 3.1 HYPOTHESES LANGUAGE

Suppose $\phi$ is any state-formula in the situation calculus whose only free variable is of sort situation[5]. Here powers of $\phi$, for example $\phi^i$, indicate that post-action formula $\phi^{i-1}$ is related by an application of regression to some pre-action formula $\phi^i$. More explicitly, $\phi^i(s) \equiv \exists \vec{x}\ \Psi_A(\vec{x}, s) \wedge Regr(\phi^{i-1}(do(A(\vec{x}), s)))$ for some $A(\vec{x})$. Thus we have that $\phi^n$, an *n-step-to-go derivation* from $\phi^0$, corresponds to the start of a formula-trajectory of length $n + 1$ leading to $\phi^0$.

Consider the set $\{\phi_j^0\}$ consisting of the state formulae in the reward axiom case statement. We can compute $\{\phi_j^1\}$ from $\{\phi_j^0\}$ by regressing the $\phi_j^0$ over all the domain's deterministic actions. Any subset of MDP states $I \subseteq S$ that are one action application from a rewarding state, model $\bigvee_j \phi_j^1$. More usefully, a state formula characterising pre-action states for each stochastic action, can be formed by considering disjunctions over $\{\phi_j^1\}$. In a similar fashion we can encapsulate longer trajectories facilitated by stochastic actions, by computing $\{\phi_j^n\}$ for $n$ larger than 1. More specifically the set of state-formulae sufficient to encapsulate such trajectories are members of the set:

$$F^n \equiv \bigcup_{i=0...n} \{\phi_j^i\}$$

---

[5]We abstain from making the situation variable explicit where doing so is superfluous.

It follows that we shall always be able to induce a classification of state-space regions by value and/or policy using state-formulae computed by regression. This is of course provided these functions have finite range[6]. If that provision is violated, we remain able to consider relational MDPs whose object universe is finitely bounded.

Our approach is based on the fact that it is much cheaper to compute $F^n$ than to perform $n$ first-order dynamic programming steps. For instance, in blocks world, we are able to compute $F^{100}$ in under a minute. Typically, not all state formulae in $F^n$ are interesting. Only a few will be relevant to an optimal or even useful value function. We propose that the useful $\phi$s be identified using inductive learning.

### 3.2 INDUCTIVE LEARNING

In what follows we provide details of our inductive algorithm supposing it will *learn* both a generalised policy and value function as a single object. So that it can learn a policy, we record for all $\phi \in F^n$, the deterministic action from which $\phi$ was derived as well as the aggregate stochastic action of which the deterministic action forms a part.

As a starting point, we assume a set of training examples. Each is triple $\eta = \langle e, v, B(\vec{t}) \rangle$, where $e$ is an MDP state, $v$ is the optimal value for $e$, and $B(\vec{t})$ is the optimal ground stochastic action for $e$. The value and first-order policy prescribed by the induced function must agree with both the value and policy entry of our training examples. For the learning algorithm, we need the notion of an example *satisfying* some $\phi^i(s) \in F^n$. $\phi^i(s)$ is of the form $\exists \vec{x}\; \phi'(\vec{x}, s)$ where $\phi'(\vec{x}, s) \equiv \Psi_A(\vec{x}, s) \land Regr(\phi^{i-1}(do(A(\vec{x}), s)))$. We say that a training example $\eta = \langle e, v, B(\vec{t}) \rangle$ satisfies $\phi^i(s)$ iff $B$ is the composite stochastic action symbol recorded for $\phi^i(s)$ and $e$ models $\phi'(\vec{t}, s)$. This captures the intuition that the state and ground action in the example match the constraints expressed by the formula.

Initially we enlisted Alkemy[17], a generic inductive logic programming utility which learns comprehensive theories from noisy structured data. Alkemy takes as input a hypotheses language described in higher order logic and a set of examples, and is able to *greedily* induce a decision tree classifying the examples. Preliminary experiments using Alkemy demonstrated that an inductive technique holds promise. We were able to generate an encoding of $F^n$ in Alkemy's input language and let Alkemy produce a decision-tree representation of a first-order value function. For example, consider the logistics domain described e.g. in [4], where a reward is given when at least one package is in Sydney. Provided with the hypotheses language $F^4$ and about a hundred training examples, Alkemy is able to induce the optimal generalised value function shown in Table 1 in a matter of seconds. Due to the greedy nature of its search through a self imposed *incomplete* hypotheses space, in its present form Alkemy is unable to build a generalised value function for other domains that we experimented on, even when provided with an $F^n$ perfectly sufficient for that purpose.[7] This, coupled with some redundancy that was not required in our planner, led us to develop our own learner specific to a planning context.

| | |
|---|---|
| IF | $\exists b\; (Box(b) \land Bin(b, Syd))$ |
| | THEN act = NA, val = 2000 |
| ELSE | |
| IF | $\exists b \exists t\; (Box(b) \land Truck(t) \land Tin(t, Syd) \land On(b, t))$ |
| | THEN $act = unload(b, t)$, val = 1900 |
| ELSE | |
| IF | $\exists b \exists t \exists c\; (Box(b) \land Truck(t) \land City(c) \land$ |
| | $Tin(t, c) \land On(b, t) \land c \neq Syd)$ |
| | THEN $act = drive(t, Syd)$, val = 1805 |
| ELSE | |
| IF | $\exists b \exists t \exists c\; (Box(b) \land Truck(t) \land City(c) \land$ |
| | $Tin(t, c) \land Bin(b, c) \land c \neq Syd)$ |
| | THEN $act = load(b, t)$, val = 1714.75 |
| ELSE | |
| IF | $\exists b \exists t \exists c\; (Box(b) \land Truck(t) \land City(c) \land$ |
| | $\neg Tin(t, c) \land Bin(b, c))$ |
| | THEN $act = drive(t, c)$, val = 1629.01 |

Table 1: Decision-tree representation of an Alkemy policy/value function for logistics (after mild simplifications). Situation variables are omitted.

Algorithm 1 provides a listing of the pseudo code for our learner. It computes a binary tree representation of the value/policy functions where nodes correspond to regressed formulae, each connected to its children by a negative and positive arc. We can build a formula $f$ by tracing from the root node to a leaf, conjoining the formula, resp. negated formulae, at each node to $f$ depending on whether we take the positive, resp. negative, arc to that node's successors. States at the resultant leaf are those that model $f$.

There are three main aspects to our algorithm. The first is the selection of the next formula for inclusion in the decision-tree (lines 5-8), the second is the generation of the hypotheses space $F^n$ (lines 9-15), and the third is the actual construction of the tree representing the first-order policy/value function (lines 16-23).

Of these three aspects, only the first requires comment. The rest should be straightforward from the description in Al-

---

[6] Value functions and/or policies for relational MDPs may have an infinite range and require infinite representations.

[7] The authors of Alkemy are currently adding support for a complete search.

**Algorithm 1** Inductive policy construction.
1: Initialise $\{max\_n, \{\phi^0\}, F^0\}$;
2: Compute set of examples $E$
3: Call BUILD_TREE$(0, E)$

4: **function** BUILD_TREE$(n : $ integer, $E : $ examples$)$
5:   **if** PURE$(E)$ **then**
6:     **return** success_leaf
7:   **end if**
8:   $\phi \leftarrow$ good classifier in $F^n$ for $E$. **NULL** if none exists
9:   **if** $\phi \equiv$ **NULL then**
10:     $n \leftarrow n + 1$
11:     **if** $n > max\_n$ **then**
12:       **return** failure_leaf
13:     **end if**
14:     $\{\phi^n\} \leftarrow$ UPDATE_HYPOTHESES_SPACE$(\{\phi^{n-1}\})$
15:     $F^n \leftarrow \{\phi^n\} \cup F^{n-1}$
16:     **return** BUILD_TREE$(n, E)$
17:   **else**
18:     $positive \leftarrow \{\eta \in E \mid \eta$ satisfies $\phi\}$
19:     $negative \leftarrow E \backslash positive$
20:     $positive\_tree \leftarrow$ BUILD_TREE$(n, positive)$
21:     $negative\_tree \leftarrow$ BUILD_TREE$(n, negative)$
22:     **return** TREE$(\phi, positive\_tree, negative\_tree)$
23:   **end if**

gorithm 1. The algorithm starts by checking whether the examples $\eta_k = \langle e_k, v_k, B_k(\vec{t_k}) \rangle$ at the current tree node are pure (lines 5-7), that is, whether they all prescribe the same stochastic action (the $B_k$ are the same) *and* all have the same value (the $v_k$ are the same). If the examples are not pure, we select, among the state-formulae generated at this point, the $\phi$ which best discriminates between the examples (line 8). Our description in Algorithm 1 remains voluntarily non-commitment about the concrete implementation of this choice. Possibilities include accepting a $\phi$ that yields sufficient *information gain* i.e., expected reduction in entropy. Alternatively, this step could yield **NULL** until $n = max\_n$, at which point information gain or some weaker measure[8] could be used to pick up a $\phi$ from $F^{max\_n}$. The tradeoff is between prematurely adding redundant nodes to the tree and needlessly generating candidate concepts.

In any case, during the process of selection of an acceptable $\phi$, we prune from $F^n$ the formulae which are not satisfied by any example in $E$ to avoid regressing them further. In general, this may lead to removing formula-trajectories that are necessary to correctly classify training data which is not graphically connected. However, we have that such pruning is admissible when the training examples correspond to state-trajectories or policies for domain instances.

---

[8]In our implementation we use the inadmissible heuristic "number of elements in $E$ satisfying $\phi$" divided by the "number of distinct state values in the subset of $E$ which satisfy $\phi$".

### 3.3 DISCUSSION

As with other learning approaches, ours operates well given training examples in the form of state-trajectories. However, due to the nature of the state formulae present in $F^n$, we should be able to learn from far fewer state-trajectories than other learning proposals to date. We see this as an important strength of our approach. At one extreme, we can take the case of a domain for which, given training data in the form of *a single* state-trajectory, our approach is able to induce an optimal policy which generalises to *all* problem states whose values were present during training. For instance, consider a deterministic blocks world with a single action, $move(x, y)$, where we seek to minimize the number of moves to the goal. The reward case statement consists of two formulae. Say that the learner is given $F^8$ (which here consists of 16 formulae) and a single trajectory of length 8 (for instance a trajectory which changes the bottom block in a tower of 5 blocks). Suppose states along the trajectory have values ranging from 8 to 0. The learner is able to learn the optimal policy covering all instances of blocks world that are of distance 8 or less to the goal. In particular, this includes all 5 blocks instances, since 8 is the length of the longest optimal 5 blocks plan. Intuitively, the policy generated says "if we are at a distance of 8 to the goal, find two arguments to $move$ which bring us to a state at a distance of 7; Otherwise, if we are at a distance of 7 to the goal, find two arguments to $move$ which bring us to a state at a distance of 6; ..."

Moreover, our method can induce a policy which is optimal in 100% of the domain instances when there is no explicit or implicit universal quantification in the reward axiom. This is because, for domains with such rewards, optimal generalised value functions have a finite range. For instance, if our blocks world goal is to have at least 4 blocks on the table, and/or 4 free blocks, then the training data need only comprise a longest optimal trajectory to the goal/s, which is of length 3, and the hypotheses language considered need only be $F^3$. It is also important to note that generalised policies computed using our algorithm maximise expected utility thus making it more general than simple goal driven learning.

As pointed out in [23], it is often the case that the range of the value function increases with the number of objects in the domain, yielding generalised value functions with infinite range. This is typical of domains where reward relevant to the optimal policy is received when *all* objects of a given type satisfy some condition, e.g. consider the blocks world and logistics domains where we ask that all blocks be in their goal position and all packages reach their destination. A negative side of our approach is that it is still value-function driven, and as such, is not suited to infer *completely* generalised policies when the range of the generalised value function is infinite. Indeed, our learner is unable to induce a policy for states whose values were not represented in the training examples. On the same note, our approach is only suited to generating *optimal* policies, thus would never conceive approximately optimal policies such as the GN policies for blocks world [22]. Generating optimal policies may be impractical, which is why suboptimal solutions are often sought [18, 23].

Compared to purely inductive methods, our hybrid approach requires an axiomatisation of the actions in the domain. We do not feel that this is an excessive demand, especially in comparison with what a pure reasoning approach such as first-order symbolic dynamic programming would require to have any chance of being practical. Recall that the performance of symbolic dynamic programming relies on the ability of making the most out of a theorem prover to prune inconsistent formulae from the generalised value function description. We found that the theorem prover could not possibly be effective unless it was provided with the domain state constraints (e.g., constraints stating that an object cannot be two places at once, that a box is not a truck, etc ...). From our experience, axiomatising those static constrains is more time-consuming than axiomatising the domain dynamics.

Finally, one of the main motivations behind our approach was to avoid much of the most complex reasoning performed by dynamic programming. While we have succeeded in taking formula rewriting and theorem proving out of the loop, we still need model-checking to match examples with the state formulae they satisfy. Although model checking is substantially faster than automated theorem proving, it is not free. When the number of formulae remaining in $F^n$ after pruning increases rapidly with $n$, our approach becomes impractical. As shown in the experiments below, we find that model-checking is the bottleneck of our approach.

## 4 RESULTS

Our approach is substantially different from most, not the least so because it generates optimal policies and is an unorthodox combination of knowledge representation, reasoning in the form of first-order regression and induction given noise free training examples. Therefore, while papers reporting results on relational MDPs typically focus on evaluating the coverage of suboptimal generalised policy discovered by some inductive method, we find it more interesting to study the factors affecting the performance of our approach on a number of domains. Our approach is implemented in C++. The results reported were obtained on Pentium4 2.6GHz GNU/Linux 2.4.21 machine with 500MB of RAM.

Interesting characteristics of our approach which presented themselves during experimentation were apparent for both the stochastic and deterministic domains. We present results for 4 deterministic and 4 stochastic variants of the logistics and blocks world domains: LG-ALL, LG-EX, BW-ALL, BW-EX and, LG-ALL$_s$, LG-EX$_s$, BW-ALL$_s$, BW-EX$_s$ respectively. LG-ALL is a deterministic version of the logistics domain in [4] such that the reward is received where all packages are in Sydney. LG-EX is similar to LG-ALL only the reward is received where at least one package is in Sydney. The stochastic versions of these, which we have called LG-EX$_s$ and LG-ALL$_s$, have an additional property that trucks have a $0.2$ chance of mistakenly driving to the wrong city. BW-ALL is the standard blocks world domain with a single move operator, such that the reward is received when all blocks are in their goal position. BW-EX is a blocks world where the reward is received when at least three blocks are on the table. The stochastic BW domains, BW-ALL$_s$ and BW-EX$_s$, have a faulty move action, which has a $0.2$ chance of dropping the box on the table regardless of its intended destination. The only difference between these domains and those used in the deterministic and probabilistic international planning competitions are the absence of planes and the use of a single "wrong city" per drive action for logistics, and the use of a single move action rather than pickup, putdown, etc for Blocks World.

Tables 2 and 3 report results obtained during experimentation with the deterministic and stochastic domains respectively. Entries contain the name of the **domain**, the maximal depth **max_n** of regression, which we set to the number of distinct values appearing in the training data, the **size** (number of blocks, or number of boxes/trucks/cities) of the instances in the training data, the number $|E|$ of examples used, and the **type** of the training data: type P indicates that the examples have been extracted from an optimal policy covering all states of the given size — these policies where generated using the NMRDPP system [13] — while type T indicates that the examples have been extracted from optimal plans (trajectories) returned by an optimal Blocks World solver [22] for selected instances of the given size. The column labelled **time** reports the time in seconds it takes our implementation to induce a policy. This column is annotated with * if pruning was used, and with # if we interrupted the algorithm before termination. The last column reports the **scope** or applicability of the induced policy. Possible entries are either $\infty$ if our algorithm induced a completely generalised policy, or an integer $k$, in which case the policy has generalised to states whose optimal value is one of the $k$ largest values in the training set.

Rows 1 to 5 in Tables 2 and 3 demonstrate the best case behaviour of our approach. For these domains a completely generalised policy, that is a first-order policy which is optimal in 100% of the domain instances, is computed given training examples which correspond to a complete MDP policy. Even though a few examples of small size (56 examples of size 2) are sufficient to induce the generalised policy, these results are presented for progressively larger sizes and training sets so as to study the cost of model-checking. The impressive speed with which the algorithm induces a completely generalised policy is due to both the small size of the required $F^n$ and the speed with which satisfiability of elements in this set is verified against training examples. We observe that the cost of model-checking increases much faster for blocks world than for logistics. This is because in blocks world, both the formulae in $F^n$ and state model descriptions in the training data are longer.

| Domain  | max_n | size | $\|E\|$ | type | time     | scope |
|---------|-------|------|---------|------|----------|-------|
| LG-EX   | 4     | 2    | 56      | P    | 0.2      | ∞     |
| LG-EX   | 4     | 3    | 4536    | P    | 14.41    | ∞     |
| BW-EX   | 2     | 3    | 13      | P    | 0.2      | ∞     |
| BW-EX   | 2     | 4    | 73      | P    | 2.2      | ∞     |
| BW-EX   | 2     | 5    | 501     | P    | 23.5     | ∞     |
| BW-ALL  | 5     | 4    | 73      | T    | 33.9     | 5     |
| BW-ALL  | 6     | 4    | 73      | T    | 136.8    | 6     |
| BW-ALL  | 5     | 10   | 10      | T    | 131.9    | 5     |
| BW-ALL  | 6     | 10   | 10      | T    | 2558.5   | 6     |
| LG-ALL  | 8     | 2    | 56      | P    | *1.8     | 8     |
| LG-ALL  | 8     | 2    | 56      | P    | 0.5      | 8     |
| LG-ALL  | 12    | 3    | 4536    | P    | #17630.3 | 5     |
| LG-ALL  | 12    | 3    | 4536    | P    | #*263.4  | 6     |
| LG-ALL  | 12    | 3    | 4536    | P    | #*1034.2 | 9     |

Table 2: Experimental Results (Deterministic)

| Domain    | max_n | size | $\|E\|$ | type | time     | scope |
|-----------|-------|------|---------|------|----------|-------|
| LG-EX$_s$ | 5     | 2    | 56      | P    | 0.2      | ∞     |
| LG-EX$_s$ | 5     | 3    | 4536    | P    | 16.19    | ∞     |
| BW-EX$_s$ | 3     | 3    | 13      | P    | 0.3      | ∞     |
| BW-EX$_s$ | 3     | 4    | 73      | P    | 2.8      | ∞     |
| BW-EX$_s$ | 3     | 5    | 501     | P    | 29.3     | ∞     |
| BW-ALL$_s$| 4     | 4    | 73      | P    | *0.4     | 4     |
| BW-ALL$_s$| 7     | 4    | 73      | P    | *11.5    | 7     |
| BW-ALL$_s$| 8     | 4    | 73      | P    | *58.0    | 8     |
| BW-ALL$_s$| 9     | 4    | 73      | P    | *1389.6  | 9     |
| LG-ALL$_s$| 12    | 2    | 56      | P    | 2.1      | 12    |
| LG-ALL$_s$| 12    | 2    | 56      | P    | *0.7     | 12    |
| LG-ALL$_s$| 22    | 3    | 4536    | P    | #1990.8  | 12    |
| LG-ALL$_s$| 22    | 3    | 4536    | P    | #*574.4  | 14    |
| LG-ALL$_s$| 22    | 3    | 4536    | P    | #*1074.5 | 15    |

Table 3: Experimental Results (Stochastic)

The results tabulated for BW-ALL and BW-ALL$_s$ validate some of the comments we made during our discussion. We already know that a complete generalised policy is out of the question, so here we make explicit the cost of matching training examples to state formulae which they satisfy. These results show that the size of the state models in the training data, the complexity of the state formulae being considered, and the length of trajectories being considered can greatly effect the practicality of model checking. Take the deterministic case for instance, due to the cost of model-checking, we achieve better time performance with 73 examples of size 4 (a complete 4 blocks policy for a fixed goal state) than with a single 10 blocks trajectory of length 10. This is also true of LG-ALL, only in this case it is the growth in $F^n$ in addition to time consuming model checking, which contributes the most to learning time. For instance, it takes a lot longer to infer a policy with larger scope from examples of size 3 and $max\_n = 12$ than with examples of size 2 and $max\_n = 8$.

Before we conclude, we should draw the reader's attention to the last 5 entries in the tables. In particular to the time our algorithm takes to induce a policy with and without pruning. If we do not prune from $F^n$ the formulae which are not satisfied by any example in $E$, our algorithm takes considerably longer to induce the optimal $n$-stage-to-go policy for LG-ALL. This is because growth in $F^n$ without pruning, given logistic's three domain actions with lenient preconditions, is too fast for model checking to keep up.

## 5   RELATED & FUTURE WORK

Previous work in relational reinforcement learning has considered exact calculation of the optimal first-order value function using symbolic dynamic programming [4], machine learning with hand-coded background knowledge [16, 7, 5] or with a general-purpose hypotheses space for representing policies [18, 23, 10, 11], or again optimisation of a value-function approximation based on combining object-local value functions assuming that each object's local behaviour depends only on its class [6].

Of these three types of approaches, the last [6] is the least directly relevant to the work presented here. Although the language used to express regularities in the value function is rich even in comparison with those used by other relational approaches, the hypotheses space supported by the learning algorithm is rather constrained in comparison to the other learning methods. The domain specification is not exploited to the extent which the adoption of reasoning permits. In the end this is substantially different from first-order dynamic programming and to the inductive approach we propose.

Our proposal is more closely related to the first two techniques above. Taking inspiration from first-order dynamic programming, we were able to exploit first-order regression to generate a hypotheses language which is sufficient to encode optimal generalised $n$-stage-to-go policies for relational MDPs. We then proposed and experimented with a method of inducing such policies from training examples drawn from domain instances. Much of the expensive reasoning required by first-order symbolic dynamic programming is avoided by doing this. We observed that some drawbacks of our parent, in particular growth in the number and the length of regressed formulae, carry over to our technique but to a much lesser degree. Some other pitfalls remain, such as the difficulty of generalising from value functions only. Finally we demonstrated cases where our approach was able to succeed and excel in contexts where previous proposals would not.

In contrast to previous inductive approaches where reasoning about the domain dynamics is confined to the generation of suitable training data, we use reasoning to generate an hypotheses space suitable to represent the optimal $n$-stage-to-go policy. Our approach achieves a middle ground between learning techniques that rely on a general-purpose hypotheses language for representing policies [18, 23, 10, 11] and those relying on hand-coded background knowledge [16, 7, 5]. As in the former, human intervention is not required to define the hypotheses space, but as in the latter, the resulting space is targeted at the domain of interest. Finding general-purpose hypotheses languages which are expressive enough to represent good policies while restricting the attention of the learner to relevant concepts is an important challenge. We see our work as a first step towards addressing this issue. Like the majority of these approaches, our proposal learns from domain instances with a small number of objects to get a useful generalised policy and value function. Unlike such techniques, ours is not suitable where no practical representation of the optimal generalised value function or policy exists.

The most pressing item for future work is to investigate ways of mitigating poor performance in the presence of universal quantification in the reward axiom, by distancing ourselves from the purely value-driven framework. A further consideration for future work is to make use of user-provided control-knowledge to prune $F^n$ quickly during regression. Finally, we would like to experiment with the practicability of concatenating optimal $n$-stage-to-go policies induced by our approach to solve problems requiring longer planning horizons.

### Acknowledgements

Thanks to Doug Aberdeen, Joshua Cole, Bob Givan, John Lloyd, Kee Siong Ng, and John Slaney for useful discussions, and to the anonymous reviewers for their suggestions on how to improve the paper. We would like to acknowledge the support of National ICT Australia. NICTA is funded through the Australian Government's *Backing Australia's Ability* initiative, in part through the Australian Research Council.

### References

[1] A. Blum and J. Langford. Probabilistic Planning in the Graphplan Framework. In *Proc. ECP*, 1999.

[2] B. Bonet and H. Geffner. Labeled RTDP: Improving the Convergence of Real-Time Dynamic Programming. In *Proc. ICAPS*, 2003.

[3] C. Boutilier, R. Dearden, and M. Goldszmidt. Stochastic dynamic programming with factored representations. *Artificial Intelligence*, 121(1-2):49–107, 2000.

[4] C. Boutilier, R. Reiter, and B. Price. Symbolic Dynamic Programming for First-Order MDPs. In *Proc. IJCAI*, 2001.

[5] J. Cole, J.W. Lloyd, and K.S. Ng. Symbolic Learning for Adaptive Agents. In *Proc. Annual Partner Conference, Smart Internet Technology Cooperative Research Centre*, 2003. http://csl.anu.edu.au/ jwl/crc_paper.pdf

[6] C. Guestrin, D. Koller, C. Gearhart, and N. Kanodia Generalizing Plans to New Environments in Relational MDPs. In *Proc. IJCAI*, 2003.

[7] S. Dzeroski, L. De Raedt, and K. Driessens. Relational reinforcement learning. *Machine Learning*, 43:7–52, 2001.

[8] Z. Feng and E. Hansen. Symbolic LAO$^*$ Search for Factored Markov Decision Processes. In *Proc. AAAI*, 2002.

[9] N. Gardiol and L. Kaelbling. Envelope-based Planning in Relational MDPs. In *Proc. NIPS*, 2003.

[10] A. Fern, S. Yoon, and R. Givan Approximate Policy Iteration with a Policy Language Bias. In *Proc. NIPS*, 2003.

[11] A. Fern, S. Yoon, and R. Givan Learning Domain-Specific Knowledge from Random Walks. In *Proc. ICAPS*, 2004.

[12] M. Ghallab, D. Nau, and P. Traverso. *Automated Planning: Theory and Practice*. Morgann Kaufmann, 2004.

[13] Charles Gretton, David Price, and Sylvie Thiébaux. Implementation and comparison of solution methods for decision processes with non-markovian rewards. In *Proc. UAI*, 2003.

[14] E. Hansen and S. Zilberstein. LAO$^*$: A heuristic search algorithm that finds solutions with loops. *Artificial Intelligence*, 129:35–62, 2001.

[15] J. Hoey, R. St-Aubin, A. Hu, and C. Boutilier. SPUDD: Stochastic Planning using Decision Diagrams. In *Proc. UAI*, 1999.

[16] R. Khardon. Learning action strategies for planning domains. *Artificial Intelligence*, 113(1-2):125–148, 1999.

[17] J.W. Lloyd. *Logic for Learning: Learning Comprehensible Theories from Structured Data*. Springer, 2003.

[18] M. Martin and H. Geffner. Learning generalized policies in planning using concept languages. In *Proc. KR*, 2000.

[19] Mausam and D. Weld. Solving Relational MDPs with First-Order Machine Learning. In *Proc. ICAPS Workshop on Planning under Uncertainty and Incomplete Information*, 2003.

[20] W. McCune. Otter 3.3 Reference Manual. Technical Report ANL/MCS-TM-263, Argonne National Laboratory, Illinois, 2003.

[21] R. Reiter. *Knowledge in Action: Logical Foundations for Specifying and Implementing Dynamical Systems*. MIT Press, 2001.

[22] J. Slaney and S. Thiébaux. Blocks world revisited. *Artificial Intelligence*, 125:119–153, 2001.

[23] S.W. Yoon, A. Fern, and R. Givan. Inductive Policy Selection for First-Order MDPs. In *Proc. UAI*, 2002.

[24] H. Younes and M. Littman. PPDDL1.0: An extension to PDDL for Expressing Planning Domains with Probabilistic Effects, 2004. http://www.cs.cmu.edu/~lorens/papers/ppddl.pdf.